\pdfoutput=1

\documentclass[letterpaper,10pt,conference]{ieeeconf}

\IEEEoverridecommandlockouts
\usepackage{amsmath}
\usepackage{amssymb}
\usepackage{booktabs}
\usepackage{graphicx}
\usepackage{url}

\title{\LARGE \bf
Learning Structured Locomotion References for Optimization-Based Whole-Body Control
}

\author{\"Omer Tar\i k Banu\c{s}$^{1,2}$, Ulu\c{c} Saranl\i$^{1,3}$, and M. Mert Ankaral\i$^{1,4}$%
\thanks{$^{1}$Center for Robotics and Artificial Intelligence (ROMER), Middle East Technical University (METU), 06800 Ankara, T\"urkiye.}%
\thanks{$^{2}$Defense Systems Technologies Division, ASELSAN Inc., 06200 Ankara, T\"urkiye.}%
\thanks{$^{3}$Department of Computer Engineering, METU, 06800 Ankara, T\"urkiye.}%
\thanks{$^{4}$Department of Electrical and Electronics Engineering, METU, 06800 Ankara, T\"urkiye.}%
}

\begin{document}

\maketitle
\thispagestyle{empty}
\pagestyle{empty}

\begin{abstract}
Quadruped locomotion requires coordinating body motion, contact forces,
foothold placement, and gait timing. We present a hierarchical controller in
which reinforcement learning adapts these structured locomotion references
while a gait scheduler, reactive footstep planner, and whole-body impulse
controller retain responsibility for high-frequency execution. The model-based
backbone can sustain a nominal trot independently, allowing the policy to focus
on when and how to modify an already functional locomotion controller. We train the
complete hierarchy in a vectorized controller-in-the-loop environment using a
six-direction disturbance curriculum. Under a common simulator and evaluation
protocol, the proposed controller achieves 77.29\% mean survival across the
tested disturbance grid, compared with 50.35\% for the model-based reference
controller. Training-time channel ablations identify adaptive foothold placement
as the largest marginal contributor to grid survival and reveal
direction-dependent effects of gait-rate adaptation. Together, these results
demonstrate the viability of the proposed hierarchical controller architecture
for integrating learned adaptation with optimization-based whole-body control
through structured locomotion references.
\end{abstract}

\section{Introduction}

Quadruped locomotion is a hybrid control problem that couples continuous
floating-base dynamics with discrete changes in unilateral contact
\cite{wensing_optimization-based_2024}. The
controller must coordinate body motion, ground-reaction forces, foothold
placement, and contact timing while respecting friction and actuator limits.
During a disturbance, several of these quantities may need to change
simultaneously and within a short time.

Model-based locomotion controllers address this complexity by separating gait
scheduling, footstep planning, force generation, whole-body control, and
joint-level feedback \cite{hutter_quadrupedal_2014,bledt_mit_2018}.
Template-based controllers likewise organize locomotion through
low-dimensional gait variables \cite{saranli_template_2003}. Convex model
predictive control (MPC), for example, plans ground-reaction forces (GRFs)
using a reduced-order model \cite{di_carlo_dynamic_2018}. Kim \emph{et al.}'s
whole-body impulse control (WBIC) is a specific optimization-based whole-body
control (WBC) formulation that converts force and motion references into
dynamically consistent joint commands at high frequency
\cite{kim_highly_2019}. Related whole-body optimization methods reason more
directly about contact and full-body dynamics
\cite{neunert_whole-body_2018}. These intermediate references provide a useful
abstraction boundary: higher layers select behaviorally meaningful quantities,
while lower layers resolve full-body dynamics and joint commands. This
structure provides reliable and physically interpretable execution, but its
adaptability depends on how the references are generated. Contact sequences
and gait cadence are often
prescribed, while footholds are commonly selected using engineered rules such
as Raibert-style heuristics \cite{raibert_legged_1986}. Extending the resulting
behavior can therefore require additional modeling, planning, and manual
tuning. More general motion-optimization methods expose footholds and contact
timing as decision variables, but still require an explicit model and objective
\cite{winkler_gait_2018,jenelten_tamols_2022}.

\begin{figure}[t]
  \centering
  \includegraphics[width=\columnwidth]{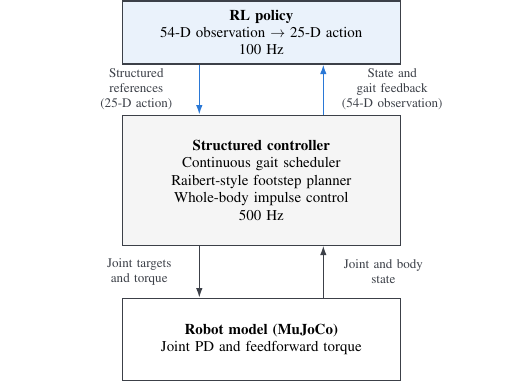}

  \caption{Proposed controller. A 100 Hz policy supplies structured locomotion
  commands
  to a gait scheduler, footstep planner, and WBIC running at 500 Hz.
  Feedback provides the 54-dimensional observation.}
  \label{fig:architecture}
\end{figure}

Reinforcement learning (RL) offers a complementary way to adapt these
decisions through interaction. Learned policies have demonstrated agile
locomotion and robust terrain traversal in simulation and on hardware
\cite{hwangbo2019learning,lee2020anymal,miki_learning_2022}.
Here, we use \emph{joint-space RL} to denote policies that output
joint-position targets or residuals tracked by joint-level feedback, as in
\cite{rudin2022learning}. Such a policy must implicitly learn how joint
commands produce coordinated body and foot motion. Training requires
substantial experience, and the policy outputs do not by themselves enforce
dynamic or contact feasibility.
Task-space and hybrid approaches instead use learning to command physically
meaningful variables while retaining a model-based execution layer
\cite{duan_taskspace_2021,tsounis_deepgait_2020,xie_glide_2023,
gangapurwala_rloc_2022}. Related approaches augment MPC or WBC using learned
corrections, optimal-control guidance, model-error compensation, or constrained
execution layers
\cite{chen_rl_augmented_mpc_2024,kim_modular_residual_2025,
kang_rl_model_based_2023,pandala_robust_predictive_2022,
wang_wholebody_constrained_2025}.

We follow this principle with a quadruped locomotion stack comprising a
continuous gait scheduler, Raibert-style footstep planning, WBIC, and
joint-level feedback, as shown in Fig.~\ref{fig:architecture}. The model-based
backbone can sustain a nominal trot from fixed motion and force references
without learned modifications. RL replaces MPC as the adaptive reference
generator rather than replacing the low-level controller. Through a
25-dimensional structured action, the policy modifies commanded body motion,
desired stance forces, foothold placement, body height, and gait progression.
The policy can therefore coordinate balance, stepping, and cadence, while the
model-based layers remain responsible for high-frequency execution.

The contribution is not the integration of RL with whole-body control alone.
Instead, we give one policy broad supervisory authority over several coupled
levels of an already functional locomotion controller. This allows learning to
focus on adapting nominal locomotion rather than rediscovering joint
coordination from scratch.

Our contributions are a hierarchical controller architecture that exposes
body motion, contact forces, foothold placement, body height, and continuous
gait timing to one learned supervisor; a vectorized MuJoCo--EnvPool environment
for training the complete controller--simulator loop; and an empirical
characterization through disturbance-survival evaluation and training-time
action-channel ablations. We evaluate the resulting configured system
under the same simulator and protocol as MPC+WBIC
\cite{kim_highly_2019} and joint-space RL \cite{rudin2022learning}. The
disturbance experiments test the resulting architecture under large
perturbations; they are not intended to establish the structured interface as
a universally optimal action representation.

\section{Proposed Hierarchical Control Architecture}

\subsection{Architecture Overview}

The controller consists of a learned policy, a gait scheduler, a reactive
footstep planner, and a whole-body impulse controller (WBIC), as shown in
Fig.~\ref{fig:architecture}. The policy sets body-motion references, desired
stance forces, foothold offsets, body height, and gait cadence. The scheduler
determines the contact sequence, the planner generates swing trajectories,
and WBIC computes joint commands from the motion and force references.

The policy runs at 100 Hz. Each action is held for five 2 ms updates of
the scheduler, planner, and WBIC, which operate at 500 Hz with state feedback.
Desired stance forces are supplied directly by the policy.

\subsection{Policy Formulation}

The policy maps an observation $\mathbf{o}_t\in\mathbb{R}^{54}$ to an action
$\mathbf{a}_t\in[-1,1]^{25}$, partitioned as
\begin{equation}
\mathbf a_t=
[\mathbf a_v^\top\;\mathbf a_f^\top\;\mathbf a_p^\top\;a_h\;a_\theta]^\top,
\end{equation}
where the components respectively command body-motion residuals, stance-force
targets, planar foothold residuals, body height, and gait progression.
Table~\ref{tab:actions} gives their physical ranges.

\begin{table}[!t]
\vspace{6pt}
\caption{Action components; leg order is front-right (FR), front-left (FL),
hind-right (HR), and hind-left (HL).}
\label{tab:actions}
\centering
\footnotesize
\setlength{\tabcolsep}{3pt}
\begin{tabular}{cll}
\toprule
Index & Command & Decoded range \\
\midrule
0--1 & body $v_x,v_y$ residual & $\pm0.5$ m/s \\
2 & yaw-rate residual & $\pm0.3$ rad/s \\
3--14 & four forces $(f_x,f_y,f_z)$ & $\pm50,\pm50,[0,250]$ N \\
15--22 & four foothold offsets $(x,y)$ & $\pm0.5$ m \\
23 & body-height residual & $\pm0.05$ m \\
24 & phase increment & $[0,0.01]$ per tick \\
\bottomrule
\end{tabular}
\end{table}

\begin{table}[!t]
\caption{Observation components.}
\label{tab:observations}
\centering
\footnotesize
\setlength{\tabcolsep}{4pt}
\begin{tabular}{lc}
\toprule
Signal & Dimension \\
\midrule
Commanded forward velocity $v_x^{\rm cmd}$ & 1 \\
Body linear velocity $\mathbf v_b^b$ & 3 \\
Body angular velocity $\boldsymbol\omega_b^b$ & 3 \\
Body orientation $\boldsymbol\phi_b$ (roll, pitch, yaw) & 3 \\
Joint positions $\mathbf q_j$ & 12 \\
Joint velocities $\dot{\mathbf q}_j$ & 12 \\
Scheduled contacts $\mathbf c$ & 4 \\
Body-frame foot positions $\{\mathbf p_{f,\ell}^b\}_{\ell=1}^4$ & 12 \\
Mean swing-phase encoding $(\sin 2\pi\bar s,\cos 2\pi\bar s)$ & 2 \\
Measured and nominal body height $(z_b,z^{\rm ref})$ & 2 \\
\midrule
Total & 54 \\
\bottomrule
\end{tabular}
\end{table}

Table~\ref{tab:observations} lists the observation channels. Actor and critic
use the same current observation, without a history buffer or privileged
disturbance information. The contact indicators describe the scheduled gait,
not measured contact.

The force channels are absolute targets; the velocity, foothold, and height
channels are residuals. The first three actions modify the external velocity
command. The resulting
horizontal velocity is integrated in world coordinates to obtain the desired
body position, with position error limited to 0.1 m. The height action
offsets a nominal stance height of 0.32 m. Desired reaction forces are applied
to stance legs, while swing legs track the planner's Cartesian trajectories.

The training reward uses bounded task-space terms
\cite{duan_taskspace_2021},
\begin{equation}
R_t=\sum_i w_i\exp(-p_i),
\label{eq:reward}
\end{equation}
with the penalties and weights in Table~\ref{tab:reward}. The set
$\mathcal A=\{0,\ldots,23\}\setminus\{5,8,11,14\}$ excludes the four vertical
force channels and the phase action from the magnitude penalty. A fall terminates training with reward
$-10$.

\begin{table*}[t]
\vspace{6pt}
\caption{Reward penalties and weights. Each healthy-state component is
$w_i\exp(-p_i)$. Superscript $w$ denotes the world frame, and $\mathbf q_0$
is the neutral-orientation quaternion.}
\label{tab:reward}
\centering
\footnotesize
\setlength{\tabcolsep}{3pt}
\begin{tabular}{@{}clc@{\hspace{13pt}}clc@{}}
\toprule
Term & Penalty $p_i$ & $w_i$ & Term & Penalty $p_i$ & $w_i$ \\
\midrule
$v_x$ & $3|v_x^w-v_x^{\rm ref}|$ & 0.25 &
Straightness & $5|y_b^w|+3|v_y^w|$ & 0.15 \\
$v_z$ & $3|v_z^w|$ & 0.10 &
Linear acceleration & $\|\mathbf a_b^w\|_2^2$ & 0.025 \\
Height & $20|z_b-z^{\rm ref}|$ & 0.35 &
Angular velocity & $\|\boldsymbol\omega_b\|_2^2$ & 0.025 \\
Orientation & $50(1-\langle\mathbf q_b,\mathbf q_0\rangle)$ & 0.15 &
Action variation & $3\|\mathbf a_t-\mathbf a_{t-1}\|_2^2$ & 0.10 \\
Phase progression & $\max(0,a_\theta+1)$ & 0.05 &
Action magnitude & $\sum_{j\in\mathcal A}a_{t,j}^2$ & 0.05 \\
\bottomrule
\end{tabular}
\end{table*}

\subsection{Continuous Gait Scheduling}

A normalized phase $\theta\in[0,1)$ and fixed per-leg offsets $o_\ell$ define
a diagonal trot. At each 500 Hz update,
\begin{align}
\theta_{k+1}&=\operatorname{mod}(\theta_k+\delta\theta_k,1),
&\delta\theta_k&=0.005(a_{\theta,k}+1),\nonumber\\
\theta_{\ell,k}&=\operatorname{mod}(\theta_k-o_\ell,1),
&c_{\ell,k}&=\mathbb I[\theta_{\ell,k}<d_\ell],
\label{eq:gait}
\end{align}
where $c_{\ell,k}$ denotes scheduled contact and $d_\ell=0.5$ is the duty
fraction. The offsets are $0$ for FR and HL and $0.5$ for FL and HR. A neutral
action gives a 0.4 s cycle; the policy can accelerate or pause progression
without changing the diagonal contact order. The scheduler also estimates
stance and swing durations for the footstep planner.

\subsection{Footstep Planning}

Foot placement follows the Raibert-style rule used by Kim \emph{et al.}
\cite{raibert_legged_1986,kim_highly_2019}. For swing progress $s_\ell$, the
remaining swing time is $t_\ell^{\rm rem}=(1-s_\ell)T_{\rm sw}$. The predicted
foothold before feedback correction is
\begin{equation}
\bar{\mathbf p}_\ell^w=\mathbf p_b^w+\mathbf R_{wb}
\left[\mathbf R_z\!\left(-\tfrac12\dot\psi^{\rm des}T_{\rm st}\right)
\mathbf p_{{\rm hip},\ell}^b+\mathbf v_{\rm cmd}^b t_\ell^{\rm rem}\right].
\label{eq:foothold_prediction}
\end{equation}
The planar Raibert correction is
\begin{equation}
\mathbf r_{\rm R}=\begin{bmatrix}
\frac12v_x^wT_{\rm st}+k_v(v_x^w-v_{x,{\rm cmd}}^w)
+\frac{z_b}{2g}v_y^w\dot\psi^{\rm des}\\
\frac12v_y^wT_{\rm st}+k_v(v_y^w-v_{y,{\rm cmd}}^w)
-\frac{z_b}{2g}v_x^w\dot\psi^{\rm des}
\end{bmatrix},
\end{equation}
and the policy modifies the touchdown location according to
\begin{equation}
\mathbf p_{\ell,xy}^{w,*}=\bar{\mathbf p}_{\ell,xy}^w+
\operatorname{clip}(\mathbf r_{\rm R},-p_{\max},p_{\max})
+\Delta\mathbf p_\ell^{\rm pol}.
\label{eq:foothold_target}
\end{equation}
Here $k_v=0.03$, $p_{\max}=0.30$ m, and clipping is componentwise. The learned
residual is added after clipping, so $p_{\max}$ bounds the heuristic rather
than the final target. The MPC+WBIC baseline uses the same heuristic bound in
its separate planner. A B\'ezier trajectory with 0.10 m clearance connects
lift-off to the target, whose world-frame height is $-0.003$ m.

\subsection{Whole-Body Control}

WBIC follows the formulation of Kim \emph{et al.}~\cite{kim_highly_2019}.
For generalized coordinates $\mathbf{q}$, contact forces $\mathbf{f}_c$,
and joint torques $\boldsymbol\tau$, the dynamics and stance constraints are
\begin{align}
\mathbf{M}(\mathbf{q})\ddot{\mathbf{q}}+
\mathbf{h}(\mathbf{q},\dot{\mathbf{q}})
 &=\mathbf{S}^{\top}\boldsymbol\tau+\mathbf{J}_c^{\top}\mathbf{f}_c,\\
\mathbf{J}_c\ddot{\mathbf{q}}+\dot{\mathbf{J}}_c\dot{\mathbf{q}}&=\mathbf{0},
\end{align}
where $\mathbf{S}$ selects actuated joints and $\mathbf{J}_c$ is the
contact Jacobian. Body and swing-foot objectives define the motion tasks.
WBIC adjusts the nominal acceleration and desired reaction forces subject
to floating-base dynamics, unilateral contact, and a friction pyramid with
coefficient $\mu=0.45$. The optimized forces can therefore differ from
the policy's desired forces.

Inverse-dynamics feedforward and joint feedback give the torque command,
\begin{equation}
\boldsymbol\tau_{\rm cmd}=\boldsymbol\tau_{\rm ff}
+\mathbf{K}_p(\mathbf{q}_j^{\rm des}-\mathbf{q}_j)
+\mathbf{K}_d(\dot{\mathbf{q}}_j^{\rm des}-\dot{\mathbf{q}}_j).
\end{equation}
Actuator limits are enforced during execution. The optimization imposes
dynamic and contact constraints on each update; it does not provide a
closed-loop stability guarantee.

\section{Training and Validation}

\subsection{Vectorized Controller-in-the-Loop Environment}

We developed a C++ EnvPool environment \cite{weng_envpool_2022} in which
each instance executes the complete controller--simulator closed loop.
MuJoCo \cite{todorov_mujoco_2012} advances the floating-base dynamics,
contacts, and actuators, while the embedded locomotion stack performs state
processing, gait and foothold generation, whole-body control, and joint-command
execution. Policy actions are held between policy updates while the 500 Hz
lower-level loop continues to run. The environment also defines initialization,
observation noise, command and disturbance generation, reward, and termination.
EnvPool parallelizes complete copies of this loop rather than replacing the
controller with a reduced training model. This design makes the proposed
hierarchical interface trainable at high throughput while preserving its
model-based execution path.

\subsection{Policy Training}

The proposed policies are trained from scratch with PPO
\cite{schulman_proximal_2017}. The primary results use
three independent policies different seeds. Each completed 85 million transitions; the
training curves plateau by this horizon, with no sustained improvement near the
end of training. The final checkpoint and its corresponding normalization
statistics are used for every evaluation condition.

Training uses 256 parallel environments and rollouts of 512 steps per
environment. Each PPO update performs ten epochs with minibatches of 1024.
The actor and critic are separate $[256,128]$ tanh networks. The learning rate
is $10^{-5}$, the entropy coefficient is 0.05, $\gamma=0.995$, and
$\lambda=0.97$. Policy and value clipping parameters are 0.1 and 0.2,
respectively; the target Kullback--Leibler (KL) divergence is 0.01, the
gradient norm is limited to 0.1, and policy standard deviations are bounded
to $[0.05,0.30]$.

The curriculum exposes the policy to smaller perturbations during initial
learning and then progressively expands the recovery task. Training disturbances are
additive velocity changes in the six signed cardinal directions. Their ceiling
increases from 1 to 2 m/s at 8.1 million transitions and to 3 m/s at
24.3 million transitions. Every 1.6 s, a disturbance is skipped with
probability 0.2; otherwise its magnitude is drawn from the upper half of the
current range with probability 0.75 and from the lower half otherwise.
Initial conditions include a $\pm0.03$ m spawn-height perturbation. Commands
are sampled per episode with $v_x\in[0,1]$ m/s,
$v_y\in[-0.2,0.2]$ m/s, and yaw rate in $[-0.2,0.2]$ rad/s.

\subsection{Validation Protocol}
\label{sec:validation}

\begin{figure}[t]
  \centering
  \vspace{6pt}
  \includegraphics[width=0.83\columnwidth]{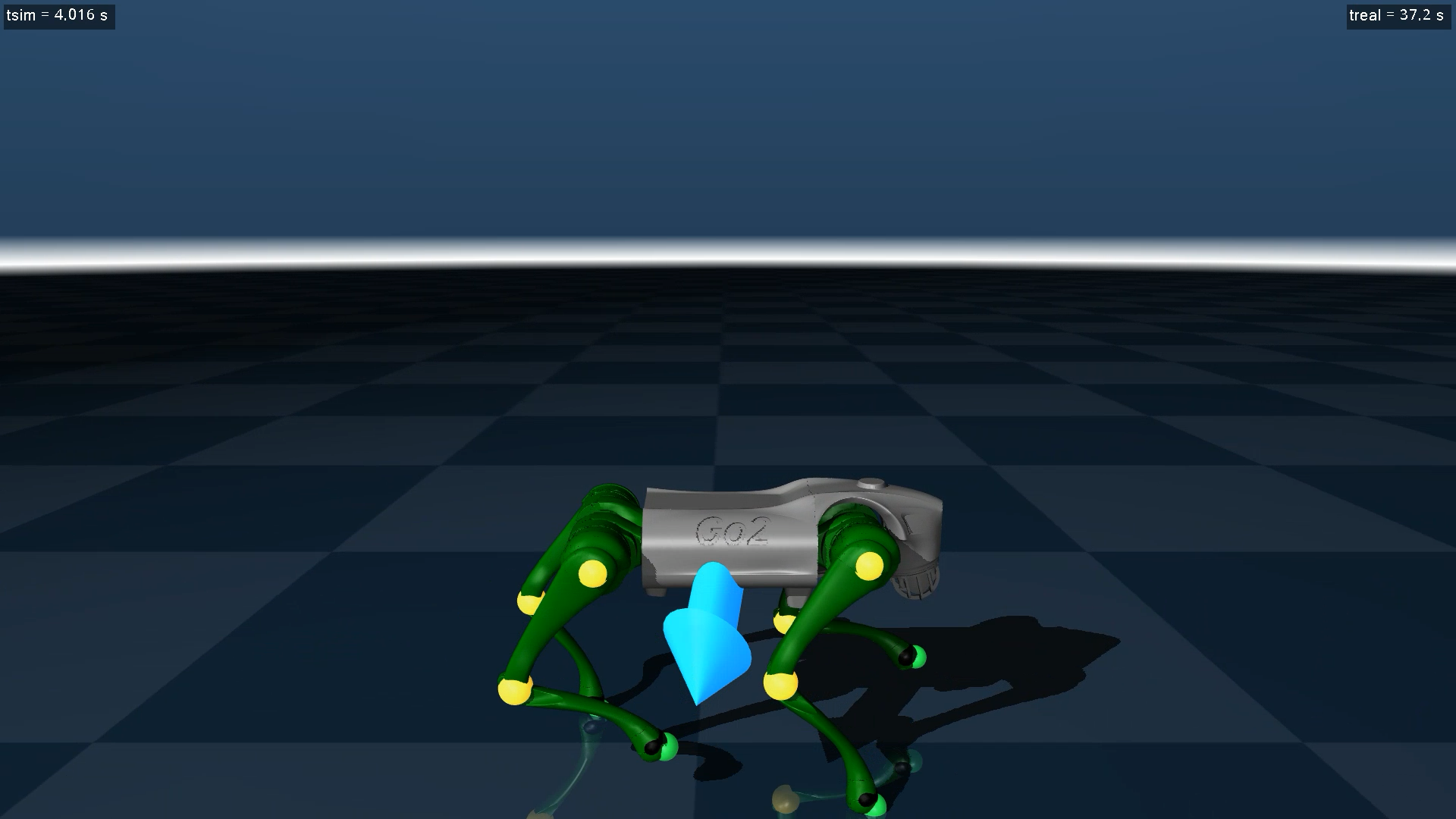}
  \caption{Go2 simulation environment. The arrow illustrates the disturbance
  direction; evaluation applies instantaneous velocity changes.}
  \label{fig:scene}
\end{figure}

All experiments use a 15 kg Go2 model on flat ground
(Fig.~\ref{fig:scene}), a 2 ms simulation step, ground friction 1.0, and joint
torque limits of $[23.7,23.7,45.43]$ N$\cdot$m for the abduction, hip, and knee
joints. Training-time friction randomization is disabled during evaluation. Uniform observation noise has
half-widths of 0.05 rad for orientation, 0.1 m/s for linear velocity,
0.2 rad/s for angular velocity, 0.01 rad for joint position, and 1.5 rad/s for
joint velocity; body position is unperturbed.

To contextualize survival, we include two reference configurations. The
model-based \textit{MPC+WBIC} controller is the Mini-Cheetah architecture
described by Kim \emph{et al.}~\cite{kim_highly_2019}, combining convex MPC
reaction-force planning \cite{di_carlo_dynamic_2018}, reactive foot placement,
and WBIC. The \textit{source-recipe-based joint-space RL} reference follows a Unitree Go2 recipe
adapted from Rudin \emph{et al.}~\cite{rudin2022learning} and implemented with
\texttt{rsl\_rl}. Its 50 Hz policy produces 12 joint-position offsets, scaled
by 0.25 rad and tracked with fixed gains $K_p=20$ and $K_d=0.5$. We enable its
existing body-height objective with the term $-30(z-0.30)^2$ after 1500
iterations with the source recipe and continue training for 500 iterations.
With 4096 environments and 24 rollout steps per iteration, this totals
196.608 million transitions for each of three policies. Both
references are reimplemented in the same MuJoCo--EnvPool framework and share
the Go2 plant, physics, actuator limits, initialization and noise models,
disturbance generator, and evaluation code with the proposed controller. They
retain their architecture-specific observations, actions, objectives, and
update rates; the comparison therefore contextualizes complete configured
systems rather than isolating their action representations.

The evaluation grid comprises six directions (forward, backward, left,
right, up, and down) and ten magnitudes from 0.5 to 5.0 m/s in 0.5 m/s
increments. Each of the 60 cells contains 100 episodes per policy.
The proposed and source-recipe-based joint-space results each cover three
independently trained policies; MPC+WBIC is one fixed model-based controller.
An additional undisturbed condition is excluded from the grid score.

As in Rudin \emph{et al.}~\cite{rudin2022learning}, a single disturbance is
applied as an instantaneous change in base velocity,
\begin{equation}
\mathbf{v}^{+}=\mathbf{v}^{-}+\mathbf{R}_{wb}(t_0)\Delta\mathbf{v}_b,
\end{equation}
where $\Delta\mathbf{v}_b$ is specified in the body frame at onset $t_0$.
The perturbation changes translational velocity without directly changing
angular velocity. Onset is sampled over gait phase after three cycles.
MPC+WBIC and Ours use their gait schedulers; joint-space RL uses an external
0.4 s
phase clock for onset sampling. Initial pose and joint angles are
randomized, and the forward command is sampled from $0.8\pm0.05$ m/s.
As a sensitivity check, we additionally evaluated the proposed controller
using constant 0.2 s external forces with impulses matched to the velocity
changes, $F\Delta t=m\Delta v$. Its survival trends were qualitatively
consistent with the instantaneous-velocity results, which we report to
maintain alignment with prior evaluation protocols.

An episode survives if it completes the 5.5 s post-disturbance window
without base contact force exceeding 1 N, absolute roll exceeding 0.8 rad,
absolute pitch exceeding 1.0 rad, or nonfinite state. These criteria
measure survival rather than return to the commanded motion.

\subsection{Training-Time Channel-Ablation Protocol}

To measure the contribution of each action group after the controller has
adapted to its absence, we train five additional policies from scratch using
the same PPO configuration, curriculum, and training horizon. For each policy,
one action group is overwritten with its neutral value before every simulator
step throughout training, and the same constraint is retained at evaluation.
The remaining action groups can therefore adapt during training instead of
encountering a channel removal only after the policy has converged.

Zero body-velocity residuals retain the external command, and zero foothold
residuals retain the heuristic foothold. Fixed height retains the nominal
0.32 m body target, while fixed cadence retains the nominal phase increment of
0.005 per controller update. The fixed-GRF condition sets $f_x=f_y=0$ and
$f_z=125$ N per stance leg. Because the GRF action supplies an absolute target
rather than a residual, this condition removes learned force modulation but
not the stance force itself. Each retrained controller and the corresponding
unablated reference are evaluated on the complete 60-cell grid with 100
episodes per cell, plus an undisturbed control condition. The common grid spans
0.5--5.0 m/s. When backward survival remains above 50\% at 5.0 m/s, evaluation
continues to the first crossing solely to estimate $c_{50}$; these extension
trials are excluded from the grid-survival score. The unablated and ablated
arms use different evaluation seeds.

\subsection{Evaluation Measures}

Survival is the primary disturbance outcome because it is defined identically for
model-based and learned controllers and is independent of their training
objectives. The direction-specific survival curves retain the response over the
complete disturbance range. The 50\% survival threshold $c_{50}$ provides a
compact, physically interpretable survival limit and is obtained by linear
interpolation at the first downward crossing of the mean curve. A curve above
50\% throughout its evaluated range is right-censored at the largest tested
magnitude; the full curves retain any later recrossings.

Let $s_{k,d,j}$ be the survival percentage for training seed $k$,
direction $d$, and magnitude $j$. The grid score is
\begin{equation}
G_k=\frac{1}{60}\sum_{d=1}^{6}\sum_{j=1}^{10}s_{k,d,j}.
\end{equation}
Each cell receives equal weight, so the score is the expected survival rate
under a uniform selection from the tested direction--magnitude conditions.
Because it depends on the chosen grid, it is reported as a secondary summary
rather than a replacement for the curves or directional thresholds. We
distinguish variation across independently trained policies from episode-level
uncertainty for a fixed policy.

Separately, five undisturbed full-trajectory trials per controller characterize
nominal tracking and mechanical effort. We report realized forward speed,
joint-torque RMS, absolute mechanical power, and mechanical cost of transport,
$\mathrm{CoT}_{\rm abs}=\int\sum_i|\tau_i\dot q_i|\,dt/(mgD_{xy})$, where
$D_{xy}$ is horizontal displacement. All controllers receive the same nominal
0.8 m/s forward command, while their mean realized speeds vary by approximately
$\pm0.1$ m/s. Electrical losses are not included.

\section{Results}

\subsection{Training Viability}

\begin{figure}[t]
  \centering
  \includegraphics[width=\columnwidth]{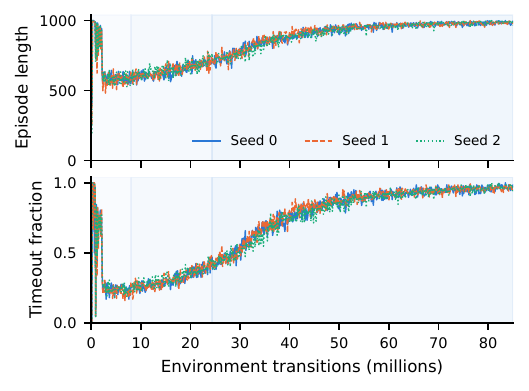}
  \caption{Training episode length and timeout fraction for three seeds.
  Shading indicates disturbance ceilings of 1, 2, and 3 m/s, with transitions
  at 8.1 and 24.3 million transitions. Curves are unsmoothed.}
  \label{fig:training}
\end{figure}

All three policies completed 85.1 million transitions without training
collapse. Figure~\ref{fig:training} shows that episode length and timeout
fraction plateau near the end of training; final timeout fractions range from
92.9\% to 98.4\%. This demonstrates that the 25-dimensional supervisory
interface can be optimized while executing the complete scheduler, planner,
and WBIC loop. 

\subsection{Disturbance Survival}
\label{sec:survival}

\begin{figure*}[t]
  \centering
  \includegraphics[width=0.98\textwidth]{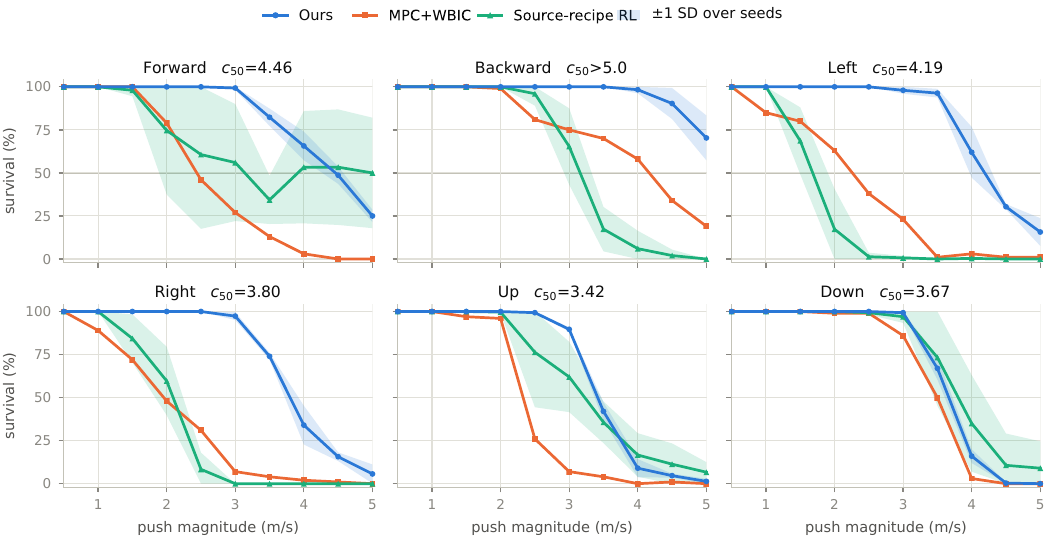}
  \caption{Survival under additive velocity disturbances, with 100 episodes
  per cell and policy. Curves and bands show the mean and $\pm1$ sample
  standard deviation across three training seeds for the proposed and
  source-recipe-based joint-space controllers. MPC+WBIC is one fixed controller.}
  \label{fig:survival}
\end{figure*}

All evaluated controller instances completed every episode in the additional
undisturbed condition, confirming that the disturbance results distinguish
disturbance survival rather than basic nominal locomotion. Under this
experimental configuration, the proposed controller achieves a mean grid
survival of $77.29\pm1.15\%$ across three training seeds, 26.94 percentage
points above the 50.35\% obtained by the model-based MPC+WBIC comparison
controller (Table~\ref{tab:comparison}). Each proposed policy survives every
episode in every direction through 2.0 m/s. For context, the source-recipe-based
joint-space configuration achieves $54.01\pm5.55\%$. This result characterizes
that configured policy rather than the attainable robustness of joint-space
RL under task-specific disturbance training.

\begin{table}[t]
  \caption{Mean survival over the 60 disturbance conditions for the complete
  configured systems. Learned-controller results report mean $\pm$ sample SD
  across three training seeds; MPC+WBIC is one fixed controller.}
  \label{tab:comparison}
  \centering
  \footnotesize
  \setlength{\tabcolsep}{7pt}
  \begin{tabular}{lr}
\toprule
Controller & Grid survival (\%) \\
\midrule
MPC+WBIC & 50.35 \\
Source-recipe-based joint-space RL & $54.01\pm5.55$ \\
Proposed & $77.29\pm1.15$ \\
\bottomrule
  \end{tabular}
\end{table}

Relative to MPC+WBIC, the proposed controller's largest threshold gains are
lateral: $c_{50}$ increases from 2.26 to 4.19 m/s leftward and from 1.96 to
3.80 m/s rightward. Backward survival remains 70.3\% at 5.0 m/s, so its
threshold lies beyond the tested range. These results demonstrate that the
proposed system has a wider disturbance envelope than the model-based
comparison controller; they do not show that structured references maximize
survival among learned control formulations. Learned-policy robustness depends
strongly on the training objective and disturbance distribution, and a
joint-space policy can likewise be optimized for this particular protocol.
The comparison therefore contextualizes complete configurations rather than
isolating the effect of the action representation.

Seed variation is concentrated near the transitions between survival and
failure. The maximum per-cell standard deviation is 14.7 percentage points at
4.0 m/s leftward, while the corresponding per-seed thresholds span only
3.95--4.26 m/s. The unablated reference used in the ablation study obtains
77.43\% grid survival, close to the 77.29\% primary mean.

\subsection{Training-Time Action-Channel Ablations}
\label{sec:ablation}

\begin{figure*}[t]
  \centering
  \includegraphics[width=\textwidth]{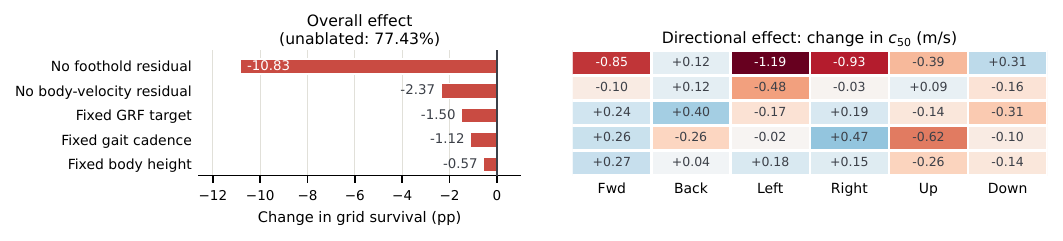}
  \caption{Training-time action-channel ablations: grid-survival change (left)
  and directional $c_{50}$ change (right; negative is worse). Backward $c_{50}$
  uses extensions above 5.0 m/s excluded from the grid score.}
  \label{fig:ablation_summary}
\end{figure*}

Figure~\ref{fig:ablation_summary} summarizes the pooled and directional
effects of training with one action group held neutral. The foothold residual
has the largest marginal contribution: without it, grid survival falls from
77.43\% to 66.60\%, a loss of 10.83 percentage points. The body-velocity,
force-modulation, cadence, and height conditions reduce the pooled score by
2.37, 1.50, 1.12, and 0.57 points, respectively. All five ablated controllers
survive every undisturbed control episode, so these differences reflect
disturbance recovery rather than failure to learn nominal locomotion.

The effects are strongly directional. Training without foothold residuals
reduces $c_{50}$ by 1.19 m/s leftward and 0.93 m/s rightward, but increases it
by 0.31 m/s downward. This sign reversal shows that the effect of foothold
adaptation is direction-dependent rather than uniformly beneficial. Although
fixing cadence changes the pooled score only modestly, it reduces upward
$c_{50}$ by 0.62 m/s and is the only ablation that reduces backward $c_{50}$
($-0.26$ m/s). The smaller height and force effects likewise change sign by
direction, so their pooled scores should not be interpreted as evidence that
the policy never uses them. The extended backward evaluation gives an
unablated $c_{50}$ of 5.15 m/s; the fixed-GRF ablation produces the largest
backward increase ($+0.40$ m/s).

\subsection{Nominal Locomotion Characteristics}
\label{sec:nominal}

\begin{figure}[t]
  \centering
  \includegraphics[width=\columnwidth]{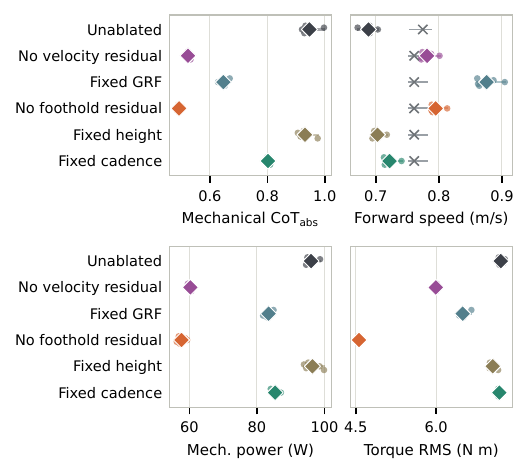}
  \caption{Nominal behavior of the retrained ablation arms over five
  undisturbed episodes. Small circles denote episodes, diamonds denote means,
  and crosses in the speed panel denote commanded-speed means.}
  \label{fig:ablation_nominal}
\end{figure}

Figure~\ref{fig:ablation_nominal} shows that the retrained fixed-cadence
controller has lower nominal mechanical power and CoT than the unablated
controller. Its pooled survival changes only modestly, with improvements in
some directions and reductions in others. Removing the foothold or
body-velocity residual likewise produces a lower-effort nominal gait. The
foothold ablation combines the lowest torque RMS with the largest grid-survival
loss, indicating that the additional actuation associated with learned
foothold correction is functional rather than uniformly inefficient.

\section{Discussion}

The results validate the hierarchy as a controller architecture in simulation.
Its distinguishing feature is the coordinated authority of one policy over
several meaningful references, while the scheduler, planner, and WBIC retain
high-frequency execution. Disturbance survival tests this interface; it does
not establish a performance-optimal form of learned control.

The ablations distinguish a channel's contribution after the controller has
adapted to its absence. Foothold adaptation is the least replaceable component:
removing it produces the only broad degradation and the largest lateral
losses. Its improvement under downward disturbances also shows that learned
foothold corrections are not uniformly beneficial and motivates
direction-dependent calibration. The fixed-cadence controller exhibits lower
nominal mechanical effort than the adaptive controller while retaining similar
pooled survival. Cadence adaptation instead changes how robustness is
distributed across disturbance directions; its observed contribution is
directional flexibility rather than demonstrated energy savings. The smaller,
signed force and height effects likewise suggest compensatory pathways rather
than uniformly inactive channels. Even
without foothold residuals, the controller retains 66.60\% grid survival, so no
single action group accounts for the complete robustness advantage. The
fixed-GRF condition remains specific to the 125 N target and does not test
removal of stance force itself.

The reference controllers contextualize validation rather than isolate action
representation. Their observations, actions, objectives, training, and update
rates differ. In particular, the lower source-recipe-based joint-space result
is not an inherent limit: learned robustness depends on training distribution,
and joint-space policies can be optimized for this benchmark. Our claim is
instead that substantial adaptation can be learned through explicit references
while model-based execution is retained. The hardware-demonstrated WBIC
architecture \cite{kim_highly_2019} provides a credible basis for future
sim-to-real work, but does not establish transfer of the complete hierarchy.

Finally, the single-channel ablations do not resolve interactions or mutual
compensation between multiple action groups, and the source-recipe-based
comparison is not training-matched. Evidence is limited to flat-ground
simulation with noisy state. An impulse-matched 0.2 s force-pulse check
produced qualitatively consistent results for the proposed controller, but was
not conducted for the reference controllers; other push durations and
rotational disturbances remain unevaluated. Survival does not establish return
to commanded motion, and transfer, safety, and hardware reliability remain
unevaluated.

\section{Conclusion}

We introduced a learned supervisor over structured locomotion references while
a scheduler, planner, and WBIC retain high-frequency execution. The hierarchy
achieves 77.29\% mean survival versus 50.35\% for the model-based reference,
and training-time ablations identify foothold adaptation as its largest
marginal channel contribution and reveal direction-dependent effects of
gait-rate adaptation. These results validate the architecture in
simulation without claiming maximum benchmark performance; broader tasks,
realistic estimation, and hardware experiments must determine its practical
advantages.

\section*{Acknowledgment}

OpenAI Codex assisted with language editing but generated no experimental
data. The authors take responsibility for the final content.

\bibliographystyle{IEEEtran}
\bibliography{references}

\end{document}